\documentclass[conference]{IEEEtran}
\usepackage{cite}
\usepackage{graphicx,epstopdf,url}
\usepackage{subfigure}
\usepackage{bbold}
\usepackage{verbatim}
\usepackage{textcomp}
\usepackage{slashbox}
\usepackage{amsmath}
\usepackage{amssymb}
\usepackage{multirow}
\usepackage{booktabs}
\usepackage[utf8]{inputenc}  
\usepackage[lined,ruled,boxed,commentsnumbered,linesnumbered]{algorithm2e}
\usepackage{indentfirst}
\usepackage{bbm}
\usepackage[all]{xy}
\usepackage{amsmath}
\DeclareMathOperator*{\argmax}{arg\,max}

\usepackage{array}
\newcolumntype{L}[1]{>{\raggedright\let\newline\\\arraybackslash\hspace{0pt}}m{#1}}
\newcolumntype{C}[1]{>{\centering\let\newline\\\arraybackslash\hspace{0pt}}m{#1}}
\newcolumntype{R}[1]{>{\raggedleft\let\newline\\\arraybackslash\hspace{0pt}}m{#1}}

\begin{document}

\title{Early-exit deep neural networks for distorted images: providing an efficient edge offloading}

\author{
    \IEEEauthorblockN{Roberto G. Pacheco\IEEEauthorrefmark{1}, Fernanda D.V.R. Oliveira\IEEEauthorrefmark{1}, Rodrigo S. Couto\IEEEauthorrefmark{1}}
   \IEEEauthorblockA{\IEEEauthorrefmark{1}Universidade Federal do Rio de Janeiro, GTA/PADS/PEE-COPPE/DEL-Poli, Rio de Janeiro, RJ, Brazil\\
    Email: pacheco@gta.ufrj.br, fernanda.dvro@poli.ufrj.br, rodrigo@gta.ufrj.br}
}

\maketitle


\begin{abstract}
Edge offloading for deep neural networks (DNNs) can be adaptive to the input's complexity by using early-exit DNNs. These DNNs have side branches throughout their architecture, allowing the inference to end earlier in the edge. The branches estimate the accuracy for a given input. If this estimated accuracy reaches a threshold, the inference ends on the edge. Otherwise, the edge offloads the inference to the cloud to process the remaining DNN layers. However, DNNs for image classification deals with distorted images, which negatively impact the branches' estimated accuracy. Consequently, the edge offloads more inferences to the cloud. This work introduces expert side branches trained on a particular distortion type to improve robustness against image distortion. The edge detects the distortion type and selects appropriate expert branches to perform the inference. This approach increases the estimated accuracy on the edge, improving the offloading decisions. We validate our proposal in a realistic scenario, in which the edge offloads DNN inference to Amazon EC2 instances.\footnote{\textcopyright2021 IEEE. Personal use of this material is permitted. Permission from IEEE must obtained for all other uses, in any current or future media, including reprinting/republishing this material for advertising or promotional purposes, creating new collective works, for resale or redistribution to servers or lists, or reuse of any copyrighted component of this work in other works.} 
\end{abstract}

\section{Introduction}
\label{sec:intro}

\begin{figure}[ht]
\center
\includegraphics[width=\linewidth]{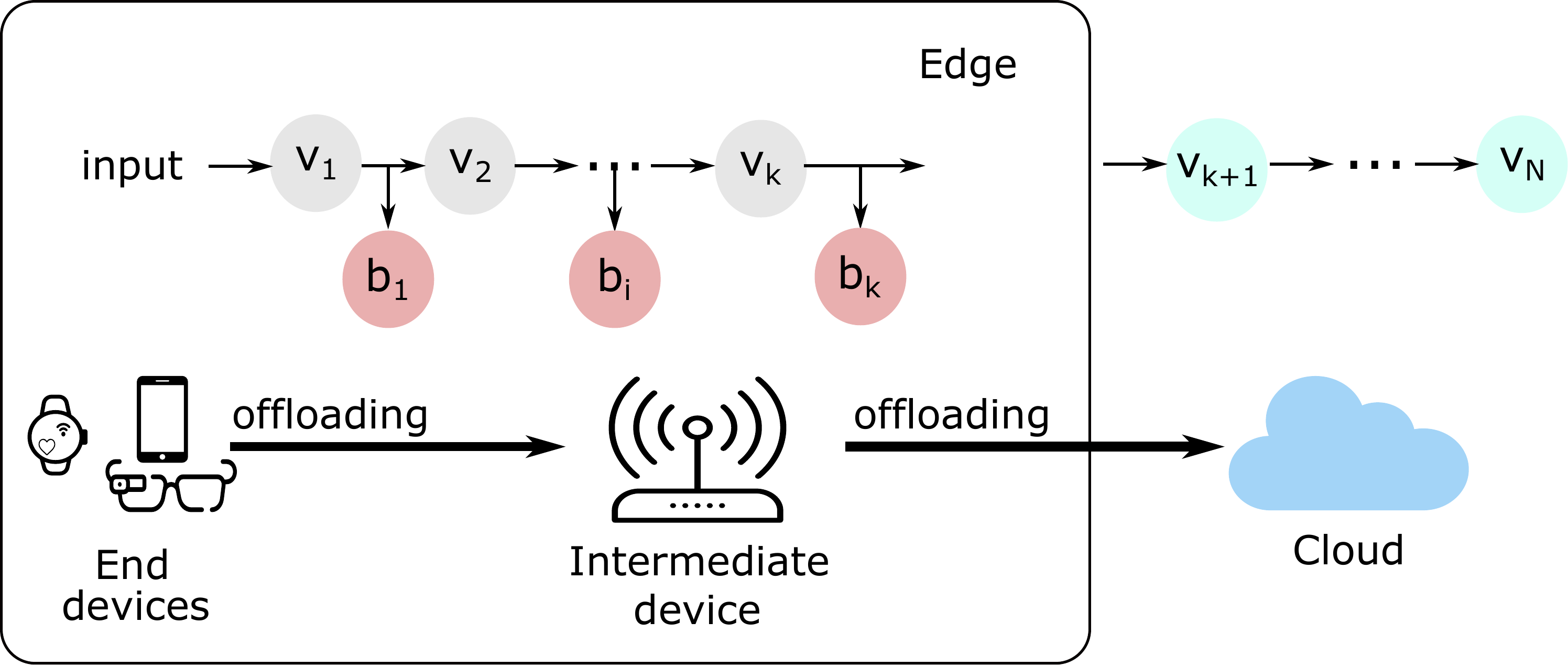}
\caption{Adaptive offloading using early-exit DNNs.}
\label{fig:adaptive_offloading}
\end{figure}

Computer vision applications employ deep neural networks (DNNs) in image classification, object detection, and many other tasks. However, DNN models require high computational power, which may not be available in end devices, such as dashcams of smart vehicles. We can overcome these constraints by offloading DNN inference to the cloud. However, cloud-based solutions are highly dependent on the network conditions, which may increase the end-to-end delay, being a barrier to low-latency applications. Edge computing emerges as an alternative to this problem, where an end device or an intermediate infrastructure (e.g., a base station) cooperates with the cloud~\cite{satyanarayanan2017emergence}. Fig.~\ref{fig:adaptive_offloading} shows an edge computing scenario with DNN partitioning, in which the DNN executes the first layers (i.e., $\text{v}_1$ to $\text{v}_k$) close to the end device and offloads the last ones to the cloud. Hence, an end device can offload computations to an intermediate device, which can offload the inference to the cloud. To ease the discussion, we call the edge the ensemble composed of end and intermediate devices. 
Also, we define edge offloading as the process of transferring the DNN inference from the intermediate device to the cloud.

This work considers an image classification application, in which a
DNN infers which object is present on an image. Also, we employ the concept of early-exit DNNs. The idea behind these DNNs is that the features extracted by the first convolutional layers may be sufficient to classify some samples with high confidence. Hence, the side branches allow the inference to be terminated at an intermediate layer, avoiding unnecessary processing until the output layer.
In this case, the offloading is adaptive since deciding where to terminate the inference depends on the input's complexity~\cite{pacheco2021calibration}.
A factor that can affect this complexity is image distortion. Cameras can gather images with various distortion types, such as blur and noise~\cite{secci2020failures}. 
The adaptive offloading scenario of Fig.~\ref{fig:adaptive_offloading} requires that the branches $\text{b}_i$ provide sufficient confident prediction for the inference to end at the edge~\cite{pacheco2021calibration}. However, Dodge and Karam~\cite{dodge2016understanding} show that DNN accuracy, and thus its confidence~\cite{pacheco2021calibration}, is sensitive to image distortion. Although they do not analyze early-exit DNNs, we may expect that image distortion can affect the branches' confidence prediction.  

As our first contribution, we show that early-exit DNNs are also sensitive to distorted images. We also propose an early-exit DNN architecture that provides an efficient edge offloading considering distorted images. To this end, we fine-tune a pre-trained early-exit DNN for each distortion type, using data augmentation.
Once trained on distorted images, the early-exit DNN should achieve robust performance on different distortion types. 
Zhou~\textit{et al.}, however, show that we need to train a model to a specific distortion type, being difficult to generalize it to other types~\cite{zhou2017classification}. To solve this issue, we propose expert branches in early-exit DNNs to improve edge offloading decisions for several distortion types.
Nevertheless, an image may contain a mixture of various distortion types. In our proposal, the edge applies a classifier~\cite{liu2020collabar} to identify the type of prevalent distortion in the image.
Then, the edge uses this information to select specific expert branches specialized in the identified distortion.
We evaluate the proposal considering two distortion types: Gaussian blur and Gaussian noise. We show that the expert branches classify more samples at the edge than an early-exit DNN trained without distorted images, improving the efficiency of edge offloading. Furthermore, we reproduce a real edge offloading scenario using Amazon EC2 (Elastic Compute Cloud) instances hosted in three different countries. Using this scenario, we show that our proposal can reduce inference latency under different networking conditions.

This paper is organized as follows. Section~\ref{sec:related_work} reviews related work. Section~\ref{sec:dnn_early_exits} overviews the adaptive offloading scenario using early-exit DNNs. Then, Section~\ref{sec:early_exit_dnn_distorted_images} presents our proposal of early-exit DNN with expert branches, and Section~\ref{sec:trainingFineTuning} describes its training. Next, Section~\ref{sec:experiments} evaluates the proposed DNN impact on an edge offloading scenario. Finally, Section~\ref{sec:conclusion} 
presents the conclusion and the next steps.

\section{Related Work}
\label{sec:related_work}
 
The related works follow three directions: early-exit DNNs, DNN model partitioning, and distortion-tolerant DNNs.

\textbf{Early-exit DNNs.} 
BranchyNet~\cite{teerapittayanon2016branchynet} consists of an early-exit DNN architecture that decides to stop the inference on a branch based on its classification entropy. In contrast, SPINN~\cite{laskaridis2020spinn} makes this decision based on the classification confidence of a given branch.
The main goal of both BranchyNet and SPINN is to reduce the inference time. 
Teerapittayanon \textit{et al.}~\cite{teerapittayanon2017distributed} propose to distribute BranchyNet in a layered architecture, composed of end devices, intermediate ones, and the cloud. Hence, each layer of this architecture can classify the images. 
Pacheco~\textit{et al.}~\cite{pacheco2021calibration}
show that early-exit DNNs can be overconfident about their prediction, providing unreliable offloading decisions. Moreover, they show that applying a simple calibration method, called Temperature Scaling, can lead to a better offloading.
None of the previous works address the impact of distorted images.
Our work uses the training scheme of BranchyNet, and the calibration method indicated in~\cite{pacheco2021calibration}. Also, we employ the early-exit decision criteria and the side-branch placement of SPINN. 

\textbf{DNN model partitioning.} Prior works study how to split a DNN model to reduce the inference time. SPINN~\cite{laskaridis2020spinn} chooses dynamically the partitioning layer (i.e., the layer in which the offloading occurs) based on the application requirements. Neurosurgeon~\cite{kang2017neurosurgeon} estimates the processing delay for each DNN layer executed at the edge or the cloud. Neurosurgeon combines this delay with networking conditions to decide how to split the DNN model. DADS~\cite{hu2019dynamic} (Dynamic Adaptive DNN Surgery) optimally partitions a DNN model, dealing with the partitioning problem as a min-cut problem. Pacheco and Couto~\cite{pacheco2020inference} optimally partition an early-exit DNN to reduce the inference time, modeling the optimization as a shortest path problem. Unlike these works, our focus is on the impact of distorted images and not on the partitioning layer choice. We thus choose a fixed partitioning layer, placed after the last side branch. However, a dynamic DNN partitioning mechanism can consider our architecture, determining the best partitioning layer according to its goals. 

\textbf{Distortion-tolerant DNNs.} Different works focus on improving the robustness of DNNs against a particular distortion type. Rozsa~\textit{et al.}~\cite{rozsa2016towards} propose a DNN training method that provides robustness specifically to noisy images. 
Dodge and Karam~\cite{dodge2018quality} propose a mixture-of-experts model, in which each expert model is fine-tuned only for a specific type of distortion. The output of the mixture-of-experts model corresponds to a weighted sum of the output from each expert DNN model, acting as an ensemble model. Next, the work determines the optimal weights for each input image.
The disadvantage of this proposal is that it needs to store an expert DNN model for each distortion type, which can be challenging to implement at the edge due to memory constraints. 
In contrast, our proposal considers a single DNN in which only the side branches are fine-tuned for specific distortion types. Hence, early-exit DNN with expert branches can scale as the number of considered types increases. 
Collabar~\cite{liu2020collabar} also fine-tunes an expert DNN model for each distortion type instead of using expert branches. Collabar has a distortion classifier to identify the distortion type in the received image and selects the most appropriate expert DNN model. Unlike our work, Collabar does not offload the inference to the cloud.
Our work uses the distortion classifier proposed by Collabar to select the most appropriate expert branches. 
The focus of \cite{liu2020collabar}, \cite{rozsa2016towards}, and \cite{dodge2018quality} is on the DNN model itself and not on its capacity to improve edge offloading decisions. 

In a nutshell, our work uses early-exits, model partitioning, and distortion-tolerant DNNs to improve edge offloading 

\section{Adaptive offloading using early-exit DNNs}
\label{sec:dnn_early_exits}
We employ an adaptive offloading mechanism, illustrated in Fig.~\ref{fig:adaptive_offloading}, that uses the same steps described in~\cite{pacheco2021calibration}. The edge receives an input and runs an early-exit DNN to infer which object the image represents. The DNN processes the input layer-by-layer until it reaches the $i$-th side branch. On this side branch, its fully connected layer generates an output vector $\mathbf{z}_{i}$. The DNN computes, from $\mathbf{z}_{i}$, the probability vector $\boldsymbol{p}_{i} = \text{softmax}(\boldsymbol{z}_{i}) \propto \exp(\boldsymbol{z}_{i})$. Each element of this vector is the probability that the object is of a particular class.

Next, the edge computes the classification confidence as $\max \boldsymbol{p}_{i}$. If this confidence level is greater or equal to the given target confidence $p_{\text{tar}}$, then the $i$-th side branch can classify the image, and the inference ends at the edge. 
Consequently, the inferred class is $\hat{y}=\argmax(\boldsymbol{p}_{i})$.  
Thus, the image is not further processed by the following layers, reducing the inference time and avoiding offloading to the cloud. If the confidence is smaller than $p_{\text{tar}}$, the next layers must process the image until it reaches the next side branch, and thus the edge repeats the procedure described before. If none of the side branches reaches $p_{\text{tar}}$, the edge sends the output of the partitioning layer to the cloud. After receiving the data, the cloud runs the remaining layers to generate a probability vector and compute the classification confidence. If this confidence does not reach $p_{\text{tar}}$, the classification uses the most confident inferred class among the branches.

\section{Early-exit DNNs for distorted images}
\label{sec:early_exit_dnn_distorted_images}

This section details the considered distortion types and our idea of early-exit DNNs with expert branches.

\subsection{Image distortion types}
\label{sec:image_distortions}
This work considers Gaussian blur and Gaussian noise as distortion types, which are described next.

\subsubsection{Gaussian blur}
\label{subsubsec:Gaussian_blur}
The image blurring can be modeled as $g(\boldsymbol{x}; \sigma_{GB}) = f(\boldsymbol{x})*h(\boldsymbol{x}; \sigma_{GB})$, where $\boldsymbol{x}=(x_{1}, x_{2})$ denotes the coordinates of a pixel, $g$ represents the blurred image, $f$ is a pristine (i.e., high-quality) image, $*$ is the convolution operator, $h$ denotes a Gaussian blur kernel, and $\sigma_{GB}$ is the kernel's standard deviation. 
We use $\sigma_{GB}$ to define the blur level, so that a higher $\sigma_{GB}$ implies a more blurred image. This work uses $\sigma_{GB}\in\{1, 2, 3, 4, 5\}$ as employed in~\cite{borkar2019deepcorrect}. For each blur level, the kernel size is given by $4\sigma_{GB}+1$~\cite{borkar2019deepcorrect}. 

\subsubsection{Gaussian noise}
\label{subsubsec:Gaussian_noise}
The noisy image can be modeled as $r(\boldsymbol{x}; \sigma_{GN}) = f(\boldsymbol{x}) + n(\boldsymbol{x}; \sigma_{GN})$, where $r$ represents the noisy image, $f$ is a pristine image, and $n$ refers to the Gaussian noise with zero-mean Gaussian distribution and standard deviation of $\sigma_{GN}$. We use $\sigma_{GN}$ to define the noise level, so that a higher $\sigma_{GN}$ implies a noisier image.
This work uses $\sigma_{GN}\in\{5, 10, 20, 30, 40\}$, as employed in~\cite{borkar2019deepcorrect}.

\subsection{Early-exit DNNs with expert branches}
\label{sec:expert_early_exit_dnns}

Our proposal is based on a fine-tuning strategy that trains a DNN model using images with a particular distortion type~\cite{dodge2018quality}. However, a camera can gather images with several distortion types. Thus, the fine-tuning strategy requires a distinct expert DNN model for each distortion type, which may not be scalable when including many different distortions. To avoid this issue, we introduce the concept of expert branches, fine-tuning only the branches to be specialized for each type of distortion. In this work, the term branches refers to the side branches at the edge and the DNN backbone's exit at the cloud. An exit point is defined as the location of DNN backbone's where we place a branch. Thus, each exit point has an expert branch for each distortion type. Our work uses a distortion classifier to identify the prevalent distortion type in an image. Then, we use this information to select the appropriate expert branches in the exit points.

\begin{figure}[ht!]
 \center
  \includegraphics[width=0.4\textwidth]{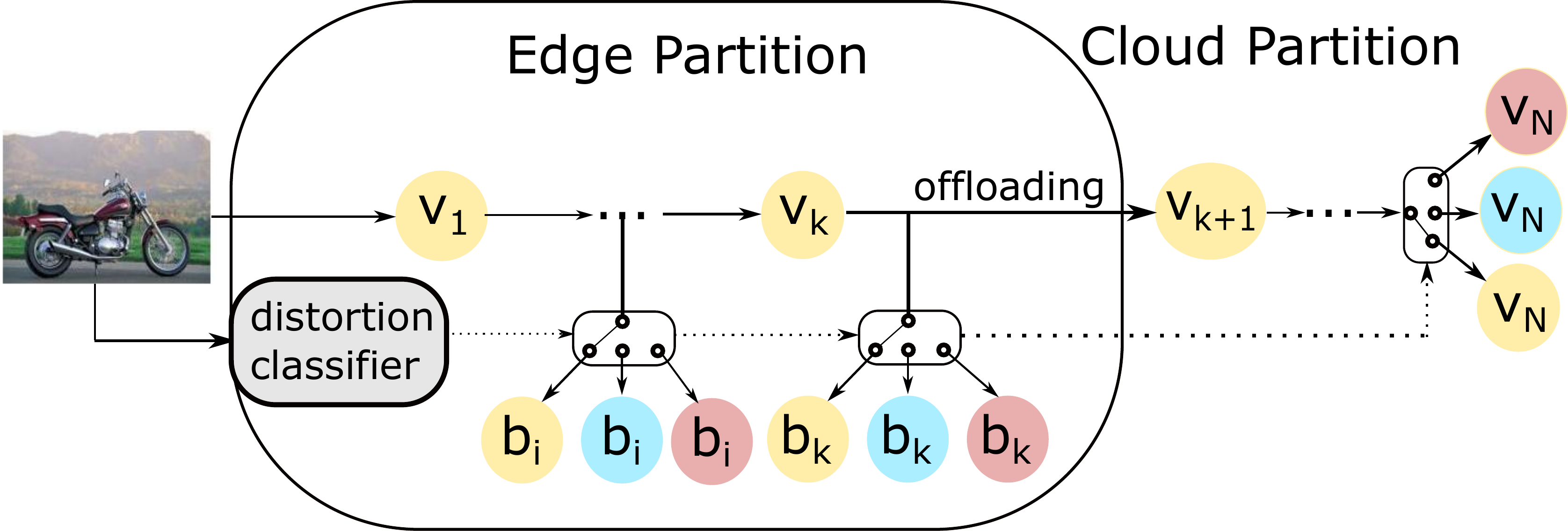}
  \caption{Early-exit DNNs with expert branches.}
    \label{fig:full_distortion_tolerant_ee_dnn}
\end{figure}

As shown in Fig.~\ref{fig:full_distortion_tolerant_ee_dnn}, our proposal is composed of a distortion classifier and an early-exit DNN with expert branches. The proposal works as follows. First, the edge receives an image.
After receiving it, the edge runs the distortion classifier, which identifies the distortion type contained in the image.
The classification chooses which expert branch is activated in the exit points, according to the distortion identified. If the classifier detects no distortion, it selects the pristine branches. After selecting the expert branches, the edge runs the inference considering only the selected branches for each exit point. 

This work uses the distortion classifier of Collabar~\cite{liu2020collabar}, a CNN (Convolutional Neural Network) trained to identify the predominant distortion in a given image. This model is trained using a Fourier spectrum of the distorted images as input. We consider only two distortion types to validate our proposal. However, the proposed architecture can be extended to consider more distortion types.
We develop the distortion classifier and early-exit DNN with expert branches on top of the Pytorch 1.7.1 framework.

\section{Training and fine-tuning}
\label{sec:trainingFineTuning}
We use the MobileNetV2~\cite{howard2017mobilenets} architecture as our DNN backbone. It is a lightweight model and optimized for running in resource-constrained hardware~\cite{howard2017mobilenets}, such as edge devices. Also, we insert three exit points along the MobileNetV2, following the strategy proposed by SPINN~\cite{laskaridis2020spinn}. 
In this work, we use the Caltech-256 dataset~\cite{griffin2007caltech}, which contains pristine images of ordinary objects, such as motorbikes and shoes. This dataset has 257 classes. We split the dataset into 80\% of images for training, 10\% for validation, and testing.

We train our DNN as follows.
We first replace the DNN backbone's fully connected layer with a new one whose number of neurons corresponds to the number of classes in the dataset. Next, we insert the exit points through the backbone. These points have branches with a fully connected layer with the same number of neurons as before. Next, we initialize all these layers through Xavier initialization~\cite{glorot2010understanding}, while all the other layers
use the weights trained on the ImageNet dataset, provided by torchvision 0.8.2.

We first train the DNN considering that each exit point has only one side branch, which is expert in pristine images. This first training follows the traditional methodology for early-exit DNNs~\cite{teerapittayanon2016branchynet}, using only pristine images from the dataset. The training uses the Adam optimization algorithm~\cite{kingma2014adam}. The learning rate of the fully-connected layers is $0.01$. The other layers, pre-trained using ImageNet, have a learning rate of $0.0015$. 
This rate is low because we do not want to deviate so far from the pre-trained weights on ImageNet, since we consider that this initialization already provides a good starting point. 
We use mini-batches with 32 images in each, weight decay of $0.0005$ and the Cosine Annealing~\cite{loshchilov2017decoupled} method as a learning rate schedule. 
We train the model until the validation loss stops decreasing for ten epochs in a row. Once trained, we obtain the early-exit DNN with expert branches for pristine images denoted by $\mathcal{E}_{\text{pristine}}$.

Next, we train, using fine-tuning, the early-exit DNNs with expert branches $\mathcal{E}_{\text{blur}}$ and $\mathcal{E}_{\text{noise}}$, for blurred and noisy images, respectively.
We train each DNN independently. That is, for each DNN, we apply the respective distortion to the training set. Using this modified set, we train the DNN considering that each exit point has only the expert branch on the considered type.   
In each mini-batch of the training set, we apply the considered distortion type to half (i.e., 16) of the images~\cite{dodge2018quality}. 
When applying distortion to each image, we randomly choose the distortion level using a discrete uniform distribution. We choose the level as $\sigma_{GB}\sim \mathcal{U}(1, 5)$, when training $\mathcal{E}_{\text{blur}}$. Otherwise, we choose $\sigma_{GN}\sim \mathcal{U}(5, 40)$, when training $\mathcal{E}_{\text{noise}}$.    
The training of $\mathcal{E}_{\text{blur}}$ and $\mathcal{E}_{\text{noise}}$ uses the same methodology employed in $\mathcal{E}_{\text{pristine}}$. However, for each DNN, we freeze the DNN backbone's parameters with the values obtained in the $\mathcal{E}_{\text{pristine}}$ training, and we train only the branches.
After training, we calibrate the branches, following the methodology of~\cite{pacheco2021calibration}. Then, for each DNN, we calibrate the branches using the validation set for the associate distortion type. 

\section{Experiments}
\label{sec:experiments}

This section evaluates our proposal considering the accuracy, offloading probability, and end-to-end latency. All the source code developed for this work is at a repository\footnote{\url{https://github.com/pachecobeto95/distortion_robust_dnns_with_early_exit}}, including the trained models of the distortion classifier, the early-exit DNN with expert branches, and the distorted datasets. 

\subsection{Accuracy}
\label{subsec:ee_performance}
We first evaluate the overall accuracy 
of each early-exit DNN with expert branches for each distortion type when receiving images with the appropriate distortion (i.e., those that have been trained) and other types. 
For each type, we apply distortion to the test dataset, varying the distortion level (i.e., $\sigma_{GB}$ and $\sigma_{GN}$), obtaining a test dataset for each distortion level. We evaluate the accuracy of the MobileNetV2 with three exit points along with its architecture considering the confidence threshold, defined in Section~\ref{sec:dnn_early_exits}, as $p_{\text{tar}}=0.8$.

\begin{figure}[!ht]
\subfigure
[Blurred images.]
{\label{fig:overall_accuracy_blur}\includegraphics[width=0.49\linewidth]{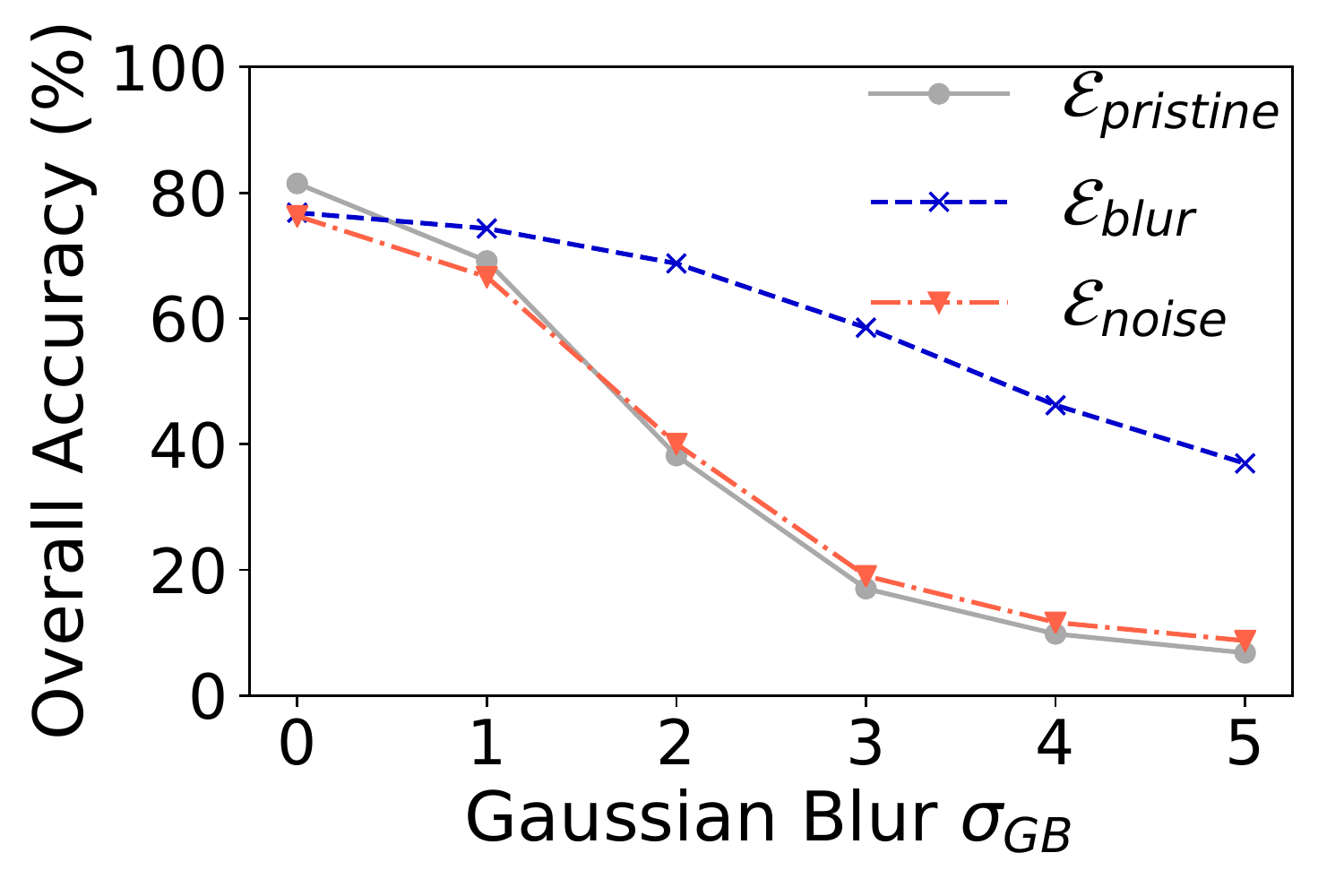}}
\subfigure
[Noisy images.]
{\label{fig:overall_accuracy_noise}\includegraphics[width=0.49\linewidth]{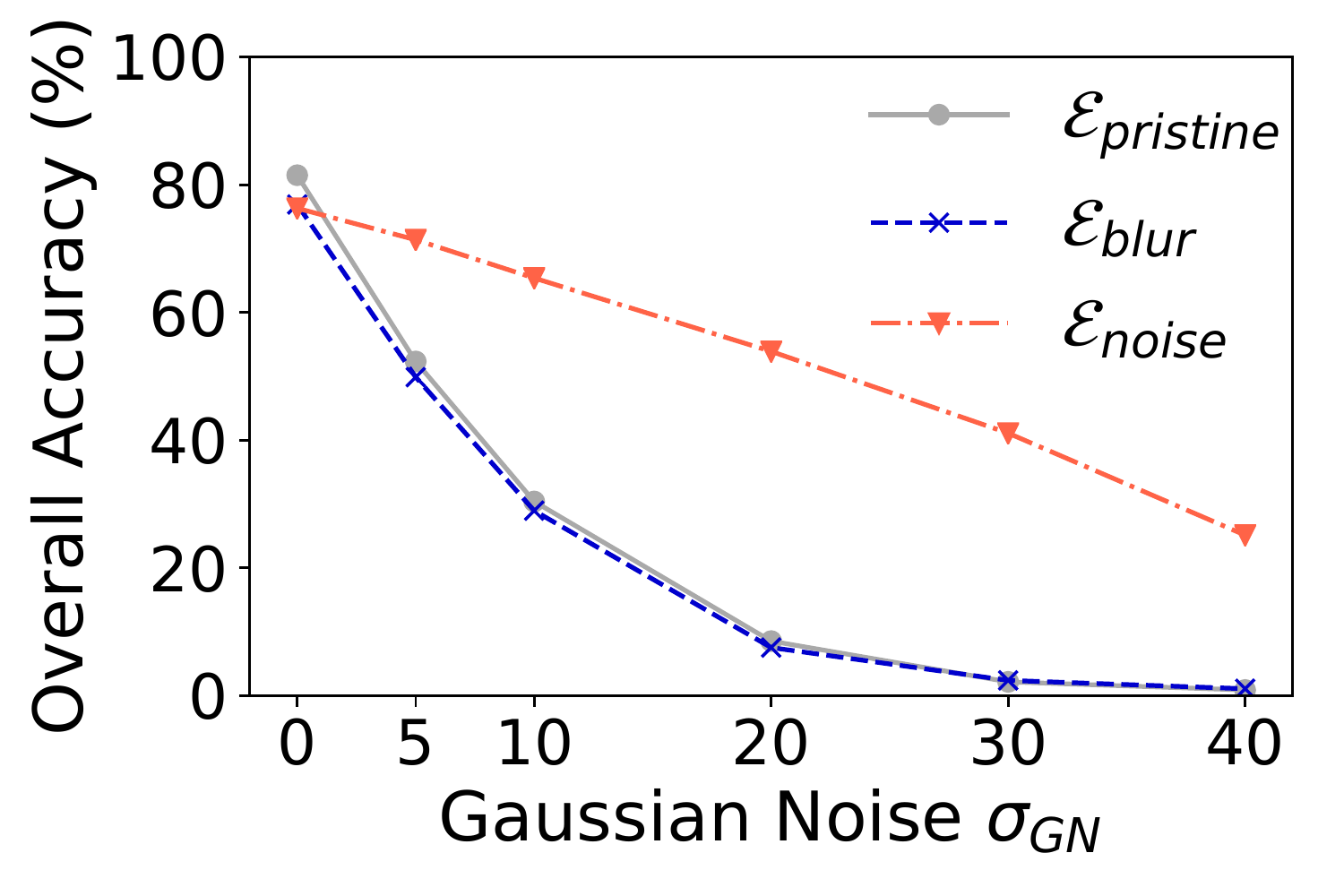}}
\caption{Overall accuracy of the early-exit DNN with expert branches.}
\label{fig:overall_accuracy_blur_noise}
\end{figure}

Fig.~\ref{fig:overall_accuracy_blur_noise} shows the overall accuracy of the early-exit DNNs versus the distortion level. Each curve in this figure corresponds to the overall accuracy of one of three DNNs (i.e., $\mathcal{E}_{\text{blur}}$, $\mathcal{E}_{\text{noise}}$, and $\mathcal{E}_{pristine}$) when executed independently without the distortion classifier. 
For example, the $\mathcal{E}_{\text{blur}}$ curve is the overall accuracy of an early-exit DNN with only blur expert branches, which was trained using blurred images.  
The overall accuracy in the y-axis is computed as the number of correct inferences divided by the total number of images in the test dataset, considering images classified on the branches at the edge and also in the cloud. In this context, the level zero of distortion corresponds to a test dataset with only pristine images. 
Moreover, the perfectly robust accuracy would be a constant curve, achieving equal accuracy at all distortion levels. 
Fig.~\ref{fig:overall_accuracy_blur} shows that, for blurred images, $\mathcal{E}_{\text{blur}}$ outperforms the other early-exit DNNs at all blur levels, except when $\sigma_{GB}=0$ (i.e., pristine images).
Unlike, $\mathcal{E}_{\text{blur}}$ performs poorly on noisy images, as shown in Fig.~\ref{fig:overall_accuracy_noise}. 
Likewise, the performance of $\mathcal{E}_{\text{noise}}$ with noisy images presents a similar behavior, as shown in Fig.~\ref{fig:overall_accuracy_blur} and~\ref{fig:overall_accuracy_noise}.
For both figures, $\mathcal{E}_{\text{pristine}}$ achieves the highest accuracy when we use only pristine images (i.e., $\sigma_{GB}=0$ and $\sigma_{GN}=0$).
These results validate our idea, demonstrating that selecting expert branches according to the distortion type 
improves the robustness of early-exit DNNs against image distortions.

\begin{figure}[!ht]
\subfigure
[Blurred images.]
{\label{fig:branch_accuracy_blur}\includegraphics[width=0.49\linewidth]{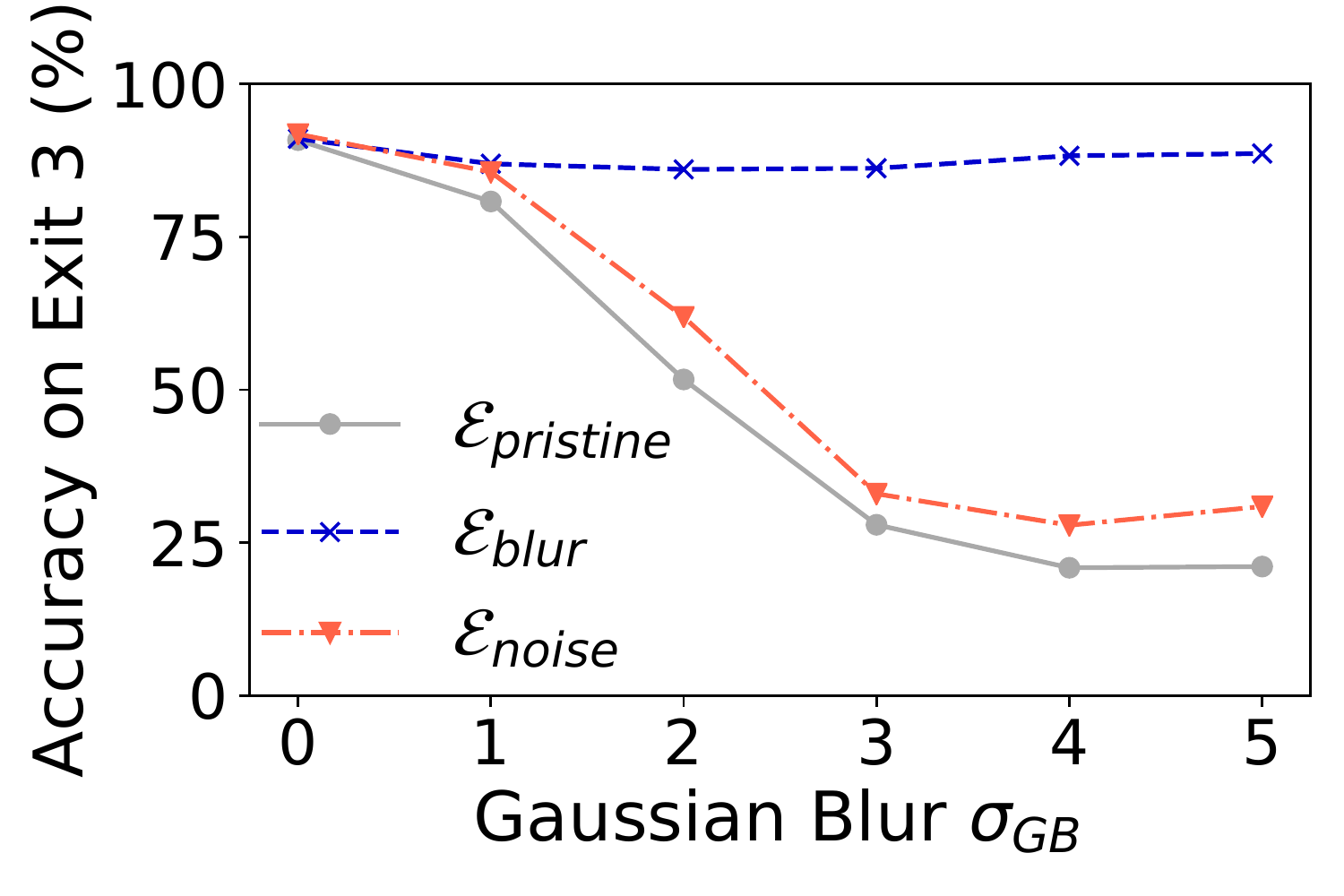}}
\subfigure
[Noisy images.]
{\label{fig:branch_accuracy_noise}\includegraphics[width=0.49\linewidth]{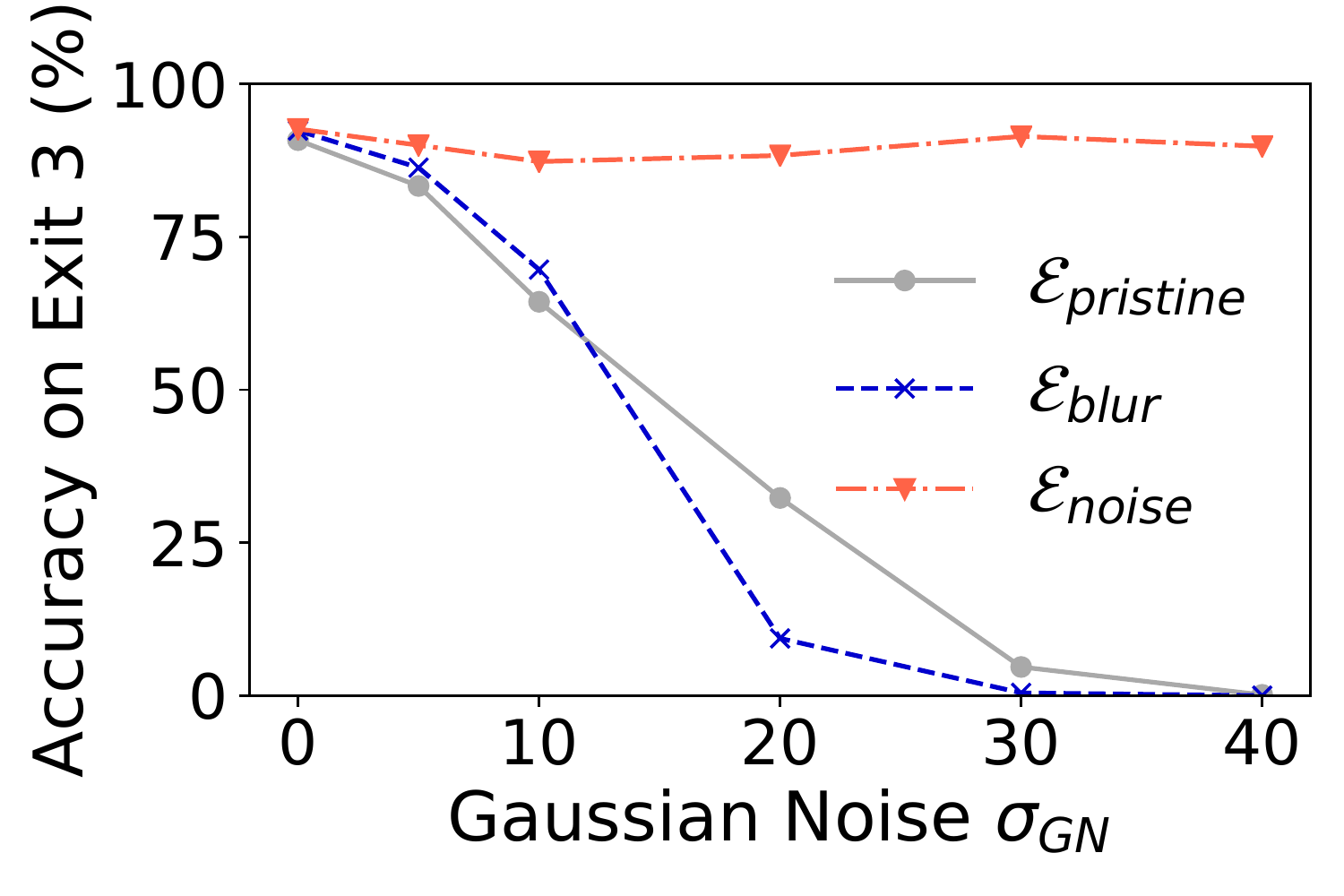}}
\caption{Accuracy on the third exit point of the expert branches.}
\label{fig:branch_accuracy_blur_noise}
\end{figure}

Next, we evaluate a particular exit point's accuracy using the same methodology as the previous experiment. This accuracy is computed as the number of correct inferences on this exit point divided by the number of samples classified on it. Fig.~\ref{fig:branch_accuracy_blur_noise} shows the accuracy on the third exit point as a function of the distortion levels. For brevity, we do not show the accuracy results for the second exit point.
Regarding the first one, it rarely can classify images, thus helping little in reducing the offloading.

In Fig.~\ref{fig:branch_accuracy_blur}, which
considers blurred images, the curve of $\mathcal{E}_{\text{blur}}$ is almost constant. Hence, this curve approaches the perfectly robust curve, as defined before. This same behavior occurs for $\mathcal{E}_{\text{noise}}$ in Fig.~\ref{fig:branch_accuracy_noise} when considering noise distortion. 
Pacheco~\textit{et al.}~\cite{pacheco2021calibration} have shown that $p_{\text{tar}}$ works as a reliable accuracy target on the device.
In both figures, the accuracy of the appropriate expert branch is always higher than the confidence threshold $p_{\text{tar}} = 0.8$ (i.e., 80\% of target accuracy). Consequently, as the side branches can reach the accuracy target, the edge does not need to send intermediate data to the cloud, improving the reliability of the offloading decisions.

\subsection{Offloading probability}
\label{subsec:offloading_probability}
We compute the probability of classifying an image on the device -- the complement of the offloading probability -- for several distortion levels, considering $p_{\text{tar}}=0.8$.
This probability is defined as the number of samples classified on the edge divided by the total number of samples in the test dataset. 
As in the previous section, we evaluate each DNN independently, without considering the distortion classifier.

\begin{figure}[!ht]
\subfigure
[Blurred images.]
{\label{fig:edge_rate_blur}\includegraphics[width=0.49\linewidth]{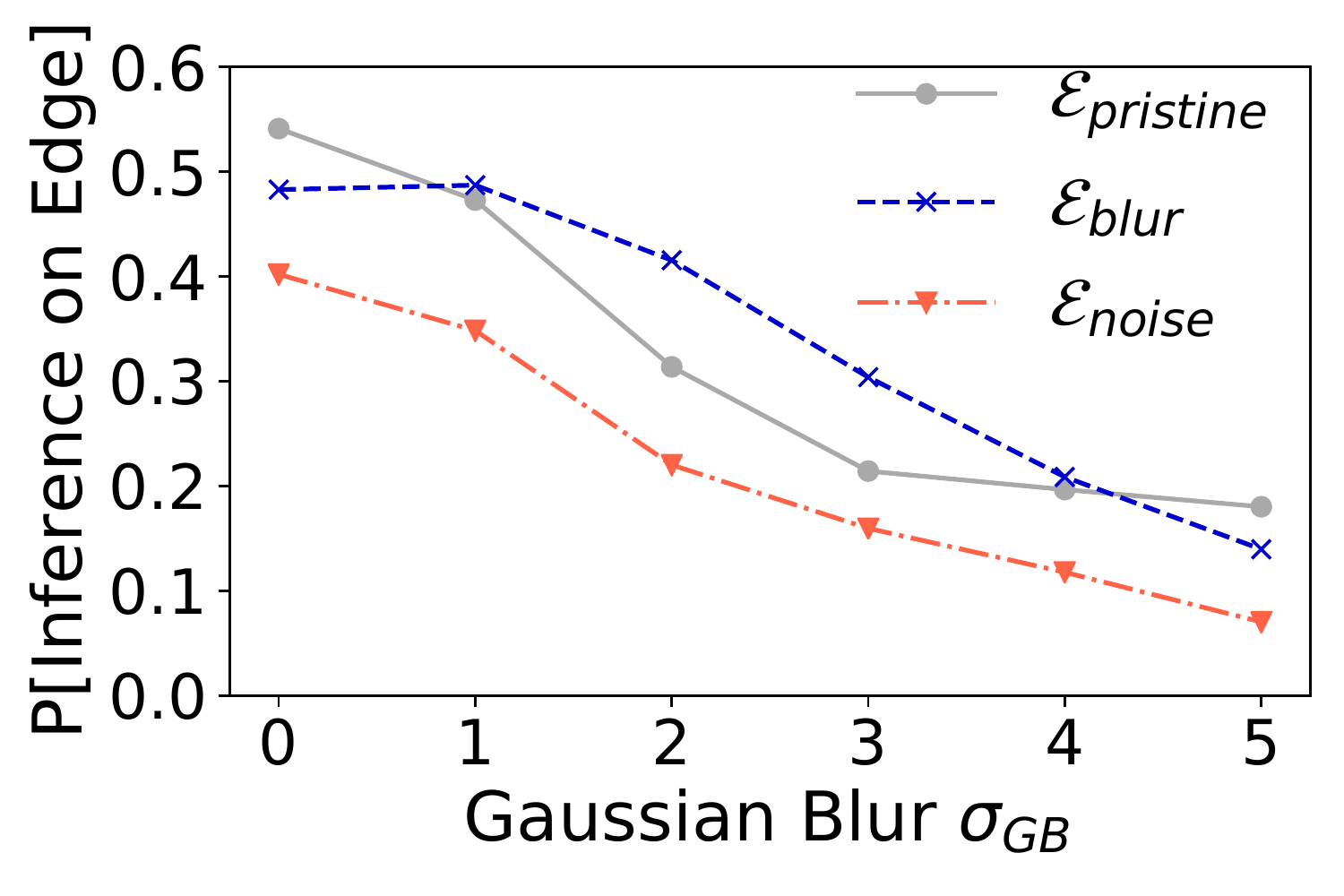}}
\subfigure
[Noisy images.]
{\label{fig:edge_rate_noise}\includegraphics[width=0.49\linewidth]{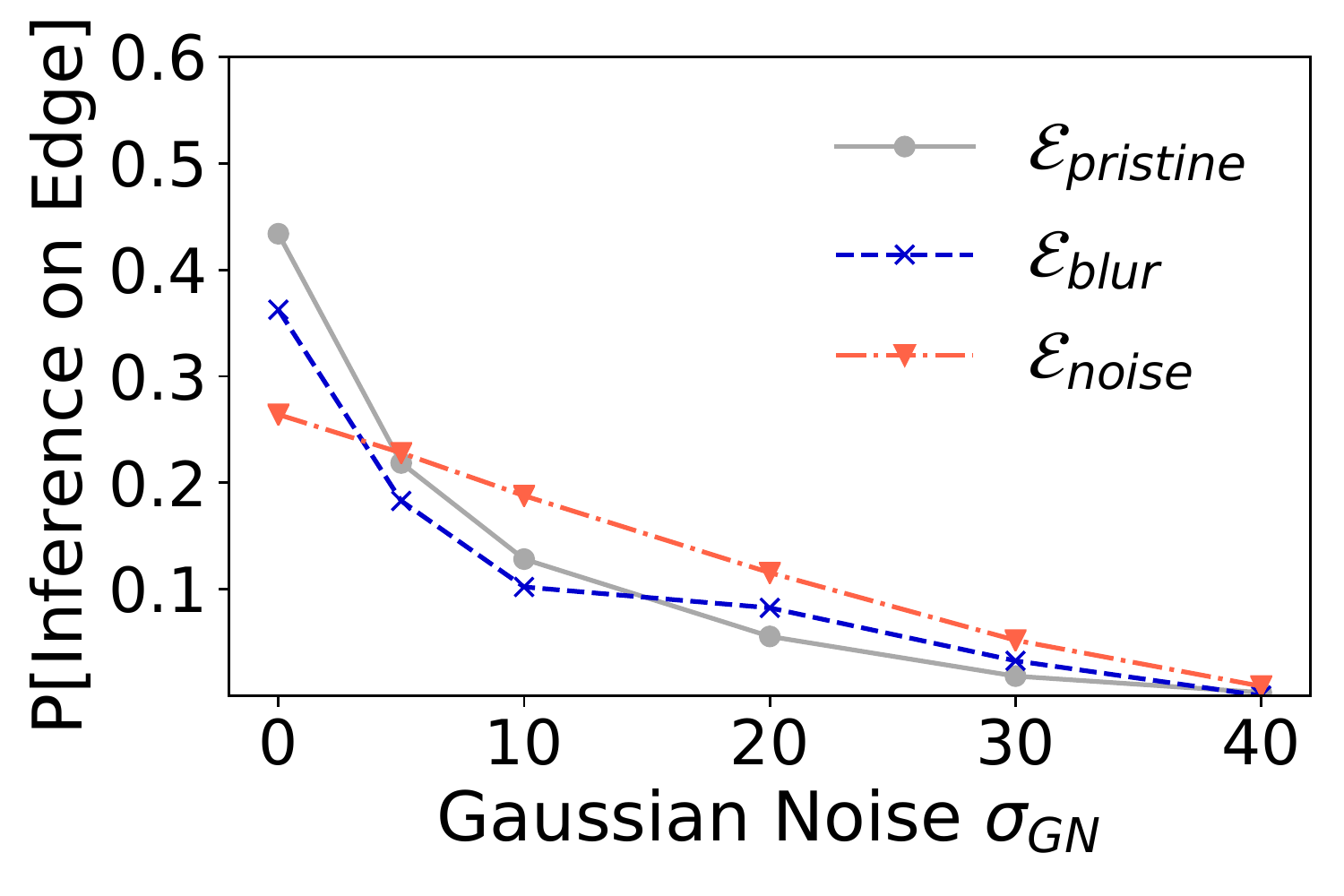}}
\caption{Probability of classifying samples on the device.}
\label{fig:edge_rate_blur_noise}
\end{figure}

Fig.~\ref{fig:edge_rate_blur_noise} shows the probability of classifying an image on the device for blurred and noisy images with several distortion levels. For blurred images, Fig.~\ref{fig:edge_rate_blur} shows that the blur expert branches $\mathcal{E}_{\text{blur}}$ can classify more images on the device, except when $\sigma_{GB} = 5$. However, in this distortion level, we can observe in Fig.~\ref{fig:overall_accuracy_blur} that the accuracy of $\mathcal{E}_{\text{pristine}}$ is much lower than $\mathcal{E}_{\text{blur}}$. Thus, despite classifying more images on the device, $\mathcal{E}_{\text{pristine}}$ also classify more samples wrongly. 
Therefore, it is better to select the most appropriate expert branches DNN, as argued by our proposal, which is the blur expert ones $\mathcal{E}_{\text{blur}}$.
Fig.~\ref{fig:edge_rate_noise} also shows that for noisy images, $\mathcal{E}_{\text{noise}}$ classifies more samples than the other experts. In the two plots of Fig.~\ref{fig:edge_rate_blur_noise}, the pristine expert can classify more images on the edge for pristine images. 
Thus, these results show that selecting the appropriate expert branches allows reaching the target accuracy on the edge more often than a traditional early-exit DNN. Hence, our proposal offloads less data to the cloud, which may reduce the end-to-end latency.

\subsection{End-to-end Latency}
\label{subsec:network_probability}
This section evaluates the end-to-end latency of Fig.~\ref{fig:full_distortion_tolerant_ee_dnn}, considering the early-exit DNN and the distortion classifier, where the edge works with the cloud to perform an image classification task.
The experiment uses a local server as the edge and an Amazon EC2 virtual machine as the cloud server to approach a real edge computing application. We implement the cloud server using Python Flask. The edge and the cloud server communicate through HTTP.
The edge server emulates the operation of the end device and the intermediate one. The edge is a bare-metal server located in Rio de Janeiro and runs a Debian 9.13 operating system.  This edge server is equipped with an Intel i5-9600K CPU with six cores at 3.70~GHz and an NVIDIA GeForce RTX 2080 Ti GPU. The cloud server is an Amazon EC2 \textit{g4dn.xlarge} instance running Ubuntu 20.04 LTS, equipped with four vCPUs from an Intel Cascade Lake CPU and an NVIDIA Tesla T4 GPU. We instantiate the cloud server in three AWS regions (i.e., geographical locations) to analyze our proposal using different network conditions.

Before running the experiment, we analyze the network conditions between the edge and the Amazon EC2 instances. We use iPerf3 and ping to obtain throughput and RTT (round-trip time) values between the edge and the cloud instance, shown in Table~\ref{tab:iperf_server}. We show these values only for illustration since the network conditions may vary during the experiment.
\begin{table}[!ht]
\caption{Network conditions for each cloud location.}
\begin{center}
\begin{tabular}{|c|c|c|c|}
\hline
AWS Region& Location& Throughput & RTT \\ \hline
sa-east-1	& São Paulo&      93 Mbps            &   12 ms          \\ \hline
us-west-1	& North California&     68 Mbps            &  182 ms          \\ \hline
eu-west-3 & Paris &      42 Mbps           &  213 ms       \\ \hline
\end{tabular}
\end{center}
\label{tab:iperf_server}
\end{table}

In this experiment, we compare our proposal with a standard early-exit DNN trained on pristine images. To do that, we measure the end-to-end latency, defined as the time between receiving the image and running the inference on the edge, or receiving the inference result from the cloud server.

\begin{figure*}[!ht]
\centering
\subfigure{\label{fig:end_to_end_blur_sp}\includegraphics[width=0.3\linewidth]{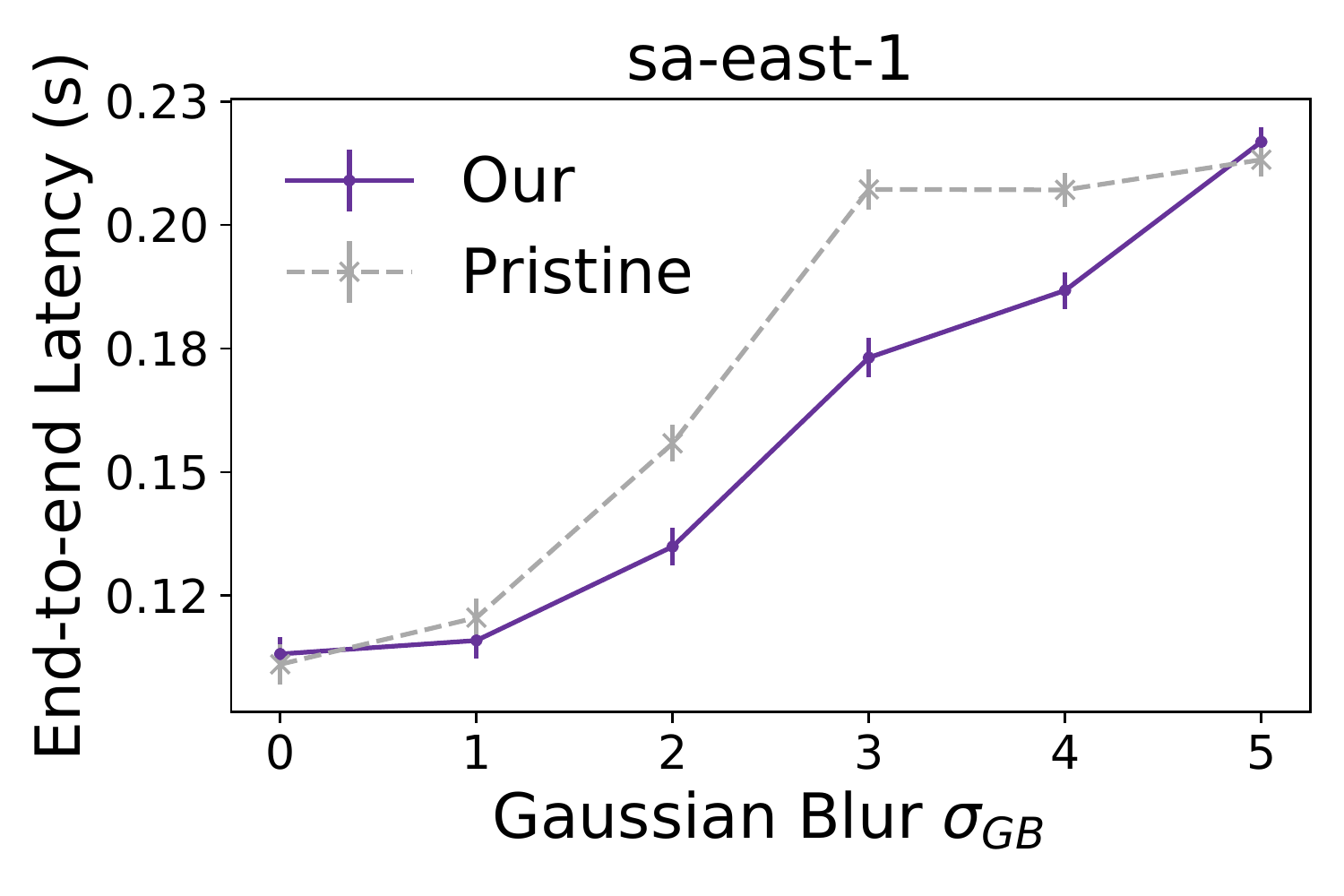}}
\subfigure{\label{fig:end_to_end_blur_fremont}\includegraphics[width=0.3\linewidth]{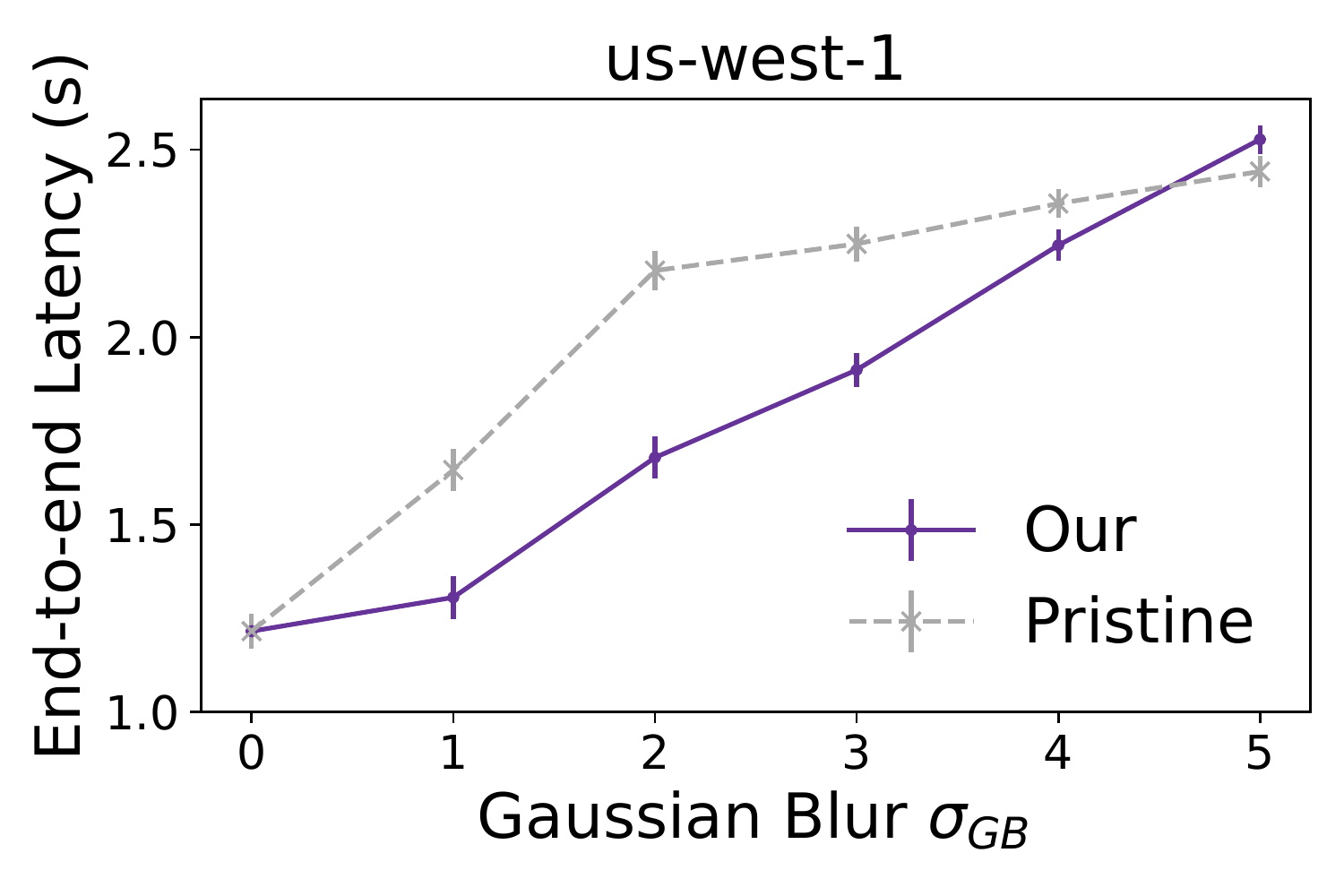}}
\subfigure{\label{fig:end_to_end_blur_paris}\includegraphics[width=0.3\linewidth]{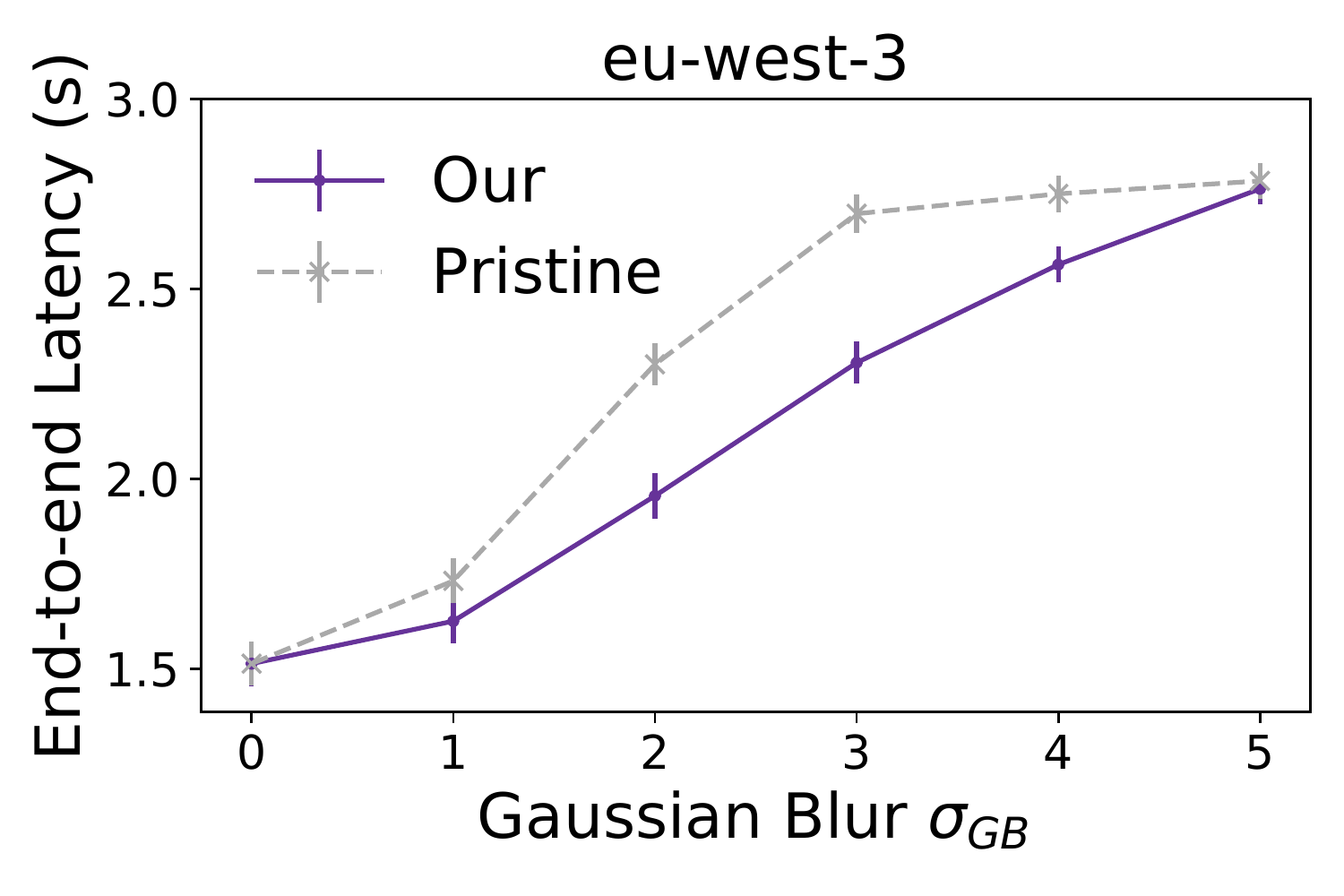}}
\subfigure{\label{fig:end_to_end_noise_sp}\includegraphics[width=0.3\linewidth]{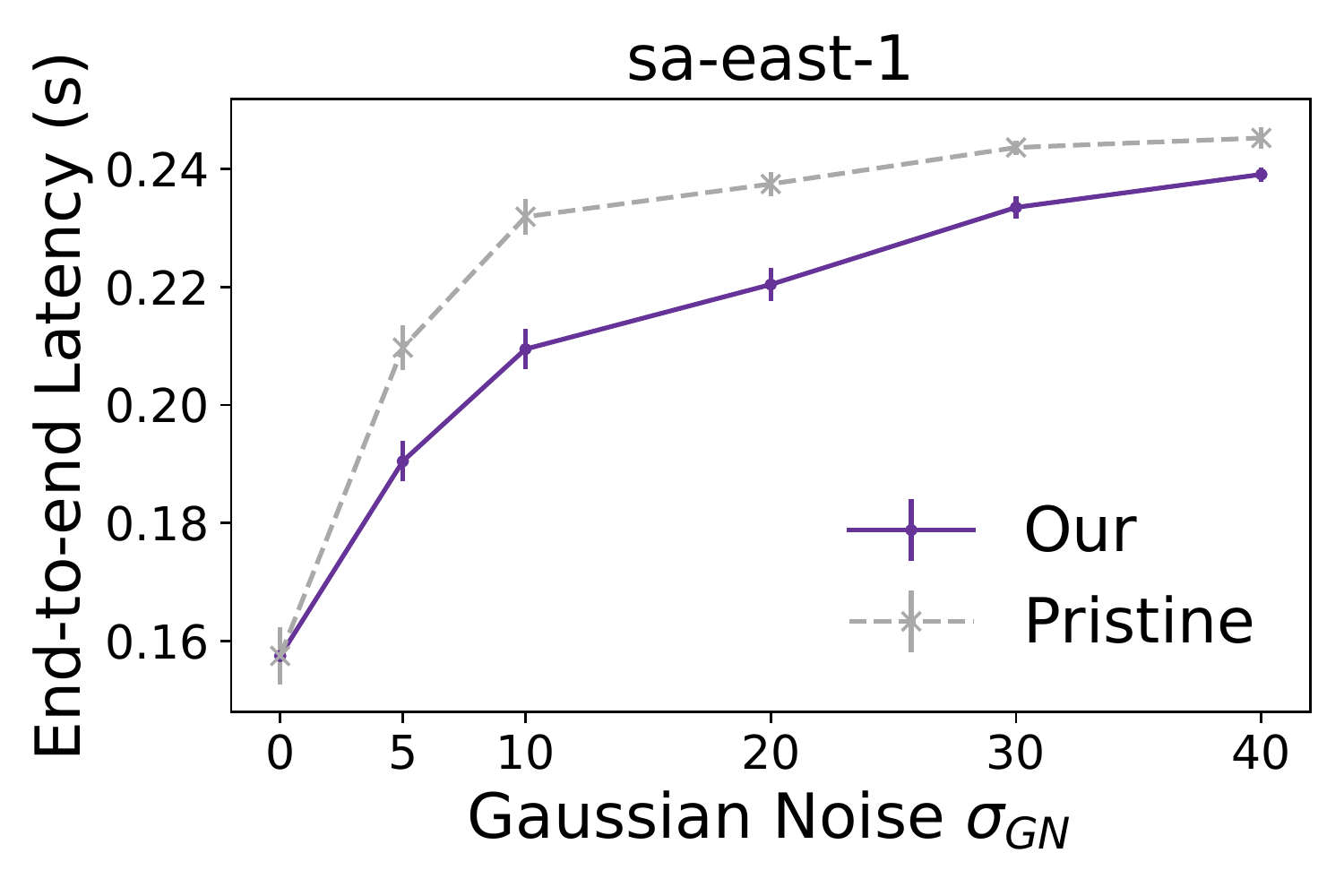}}
\subfigure{\label{fig:end_to_end_noise_fremont}\includegraphics[width=0.3\linewidth]{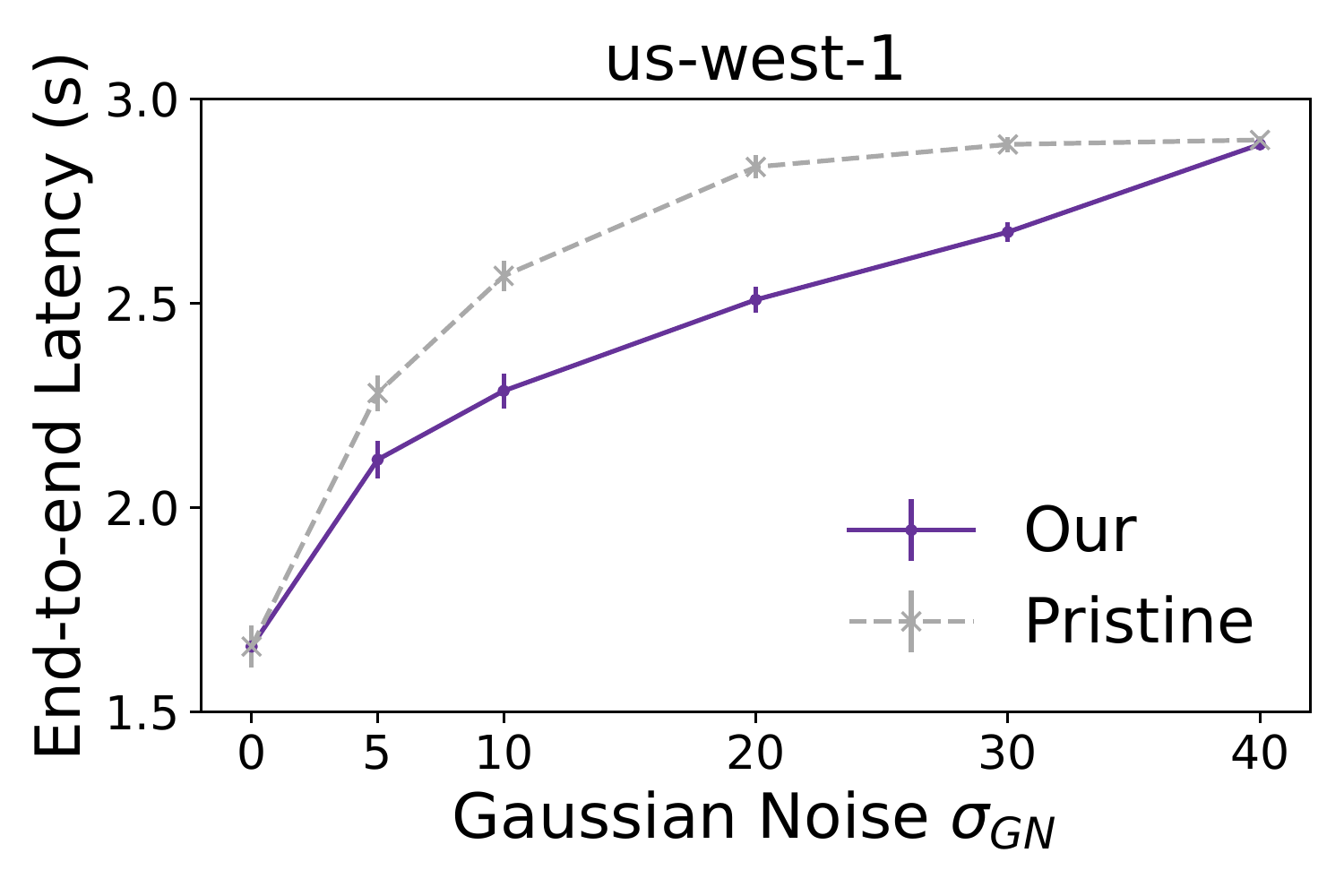}}
\subfigure{\label{fig:end_to_end_noise_paris}\includegraphics[width=0.3\linewidth]{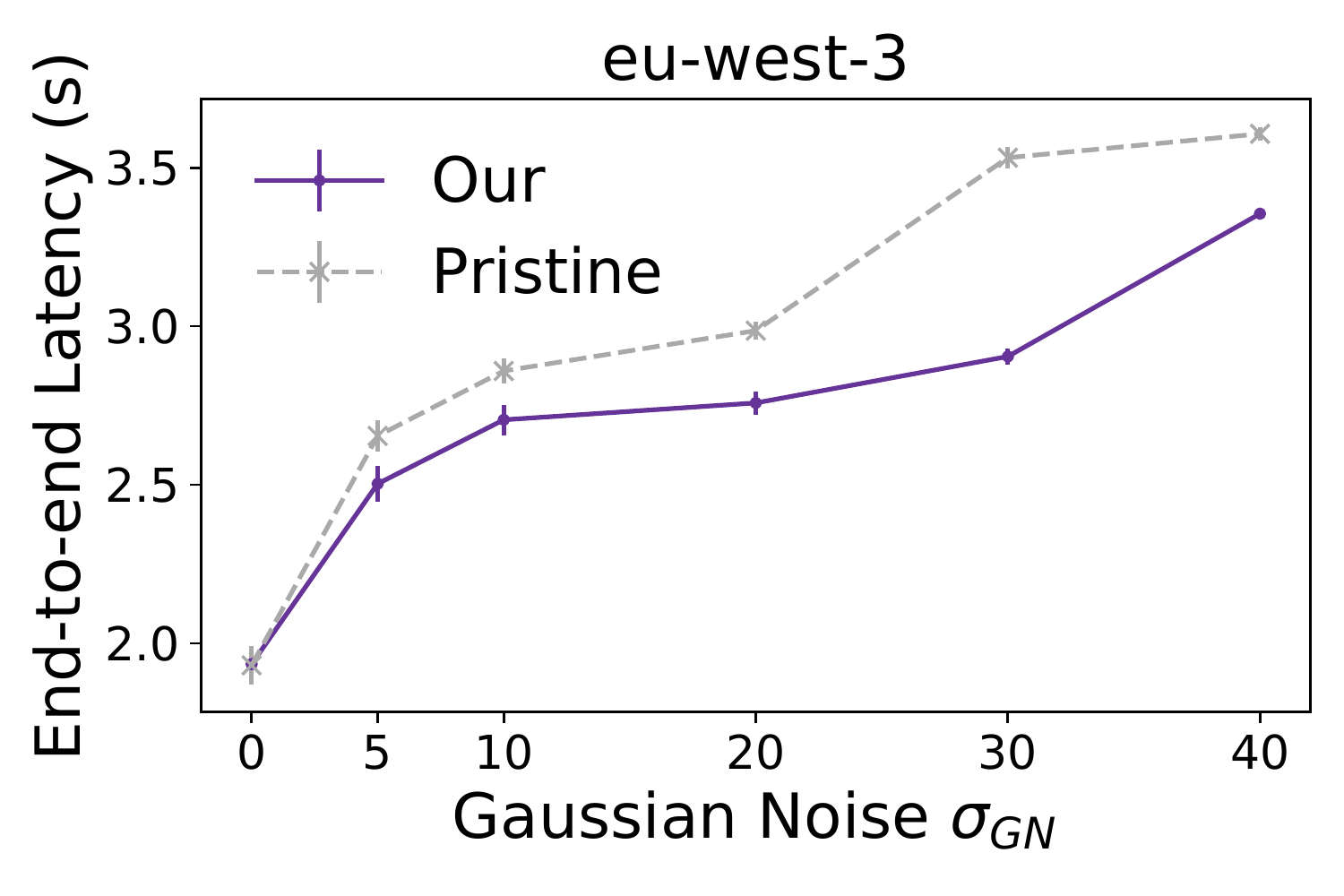}}
\caption{End-to-end latency comparison between early-exit DNN with expert branches and standard early-exit DNN trained on pristine images.}
\label{fig:end_to_end}
\end{figure*}

For each experimental run, the edge server applies a distortion type to an image.
After that, this server starts a counter to measure the end-to-end latency and runs the distortion classifier. Next, our proposal uses the distortion type identified to select the expert branches in the edge. In contrast, in the standard proposal, after applying the distortion, the edge server starts measuring the end-to-end latency and runs only the inference.  
In both cases, if the side branches can reach the confidence threshold $p_{\text{tar}}$, the inference and the counter ends. 
Otherwise, the edge server sends the intermediate results to the cloud server, which processes the remaining layers. The edge server waits for the HTTP response to finish the end-to-end latency counter. This procedure is executed for each image from the test dataset and for each AWS region. Fig.~\ref{fig:end_to_end} shows, for each region, the average end-to-end latency for the test dataset with confidence intervals of 95\%.

Fig.~\ref{fig:end_to_end} shows the end-to-end latency according to the distortion level for the three AWS regions. The top part of the figure shows the results regarding the blur, while the bottom part considers the noise. In this figure, the curves labeled as ``Our'' refer to our proposal, while the ``Pristine'' curves correspond to the standard early-exit DNN. Note that we can reduce the end-to-end latency for all locations, especially for intermediate distortion levels, such as $\sigma_{GB} = 2$, $\sigma_{GB} = 3$ or $\sigma_{GN} = 10$, $\sigma_{GN} = 20$. End-to-end latency is reduced because our proposal, through expert branches, classifies more images on the edge, as shown in Fig.~\ref{fig:edge_rate_blur_noise}, avoiding offloading to the cloud. Therefore, we can reduce the end-to-end latency for different network conditions. For example, even for cloud locations closer to the edge, such as São Paulo (i.e., the best network conditions using Table~\ref{tab:iperf_server} as reference), our proposal is still effective. Besides, our proposal increases the accuracy, as shown before.
When $\sigma_{GB} = 5$ in Fig.~\ref{fig:end_to_end}, the end-to-end latency of our proposal can exceed that of the standard early-exit DNN. As noted before, in the case of $\sigma_{GB} = 5$, the $\mathcal{E}_{\text{pristine}}$ can classify more images on the device than $\mathcal{E}_{\text{blur}}$, but with an accuracy drop. At last, note that, due to different network conditions and geographical distances, the end-to-end latency can be quite different among the cloud locations. In sa-east-1, the end-to-end latency can be less than 0.3\,s, while it can be greater than 3\,s in eu-west-3. 

At this stage, we have demonstrated that our proposal can improve DNN accuracy and reduce the end-to-end latency for different distortion types with several levels. Thus, we can make early-exit DNNs more robust against image distortion and improve offloading decisions.

\section{Conclusions}
\label{sec:conclusion}

In image classification applications, the gathered image can have several distortion types, which degrade DNN accuracy. 
This work has shown that early-exit DNN with expert branches can improve inference robustness against image distortions by increasing the DNN accuracy for various distortion levels.
Our proposal can classify more images on the edge in an adaptive offloading scenario, indicating that it can take better offloading decisions. Hence, the expert branches can reduce the end-to-end inference latency for distorted images.

As future steps, we may expand our proposal to consider branches trained on mixtures of distortions. Moreover,
a future direction is to evaluate the impact of our proposal for dynamic DNN partitioning scenarios.

\section*{Acknowledgements}
This study was financed in part by the Coordenação de Aperfeiçoamento de Pessoal de Nível Superior - Brasil (CAPES) - Finance Code 001. It was also supported by CNPq, FAPERJ Grants E-26/203.211/2017 and E-26/211.144/2019, and FAPESP Grant 15/24494-8.
\bibliographystyle{IEEEtran}
\bibliography{header,bibFile}

\end{document}